\documentclass[11pt]{article}

\usepackage[preprint]{acl}

\usepackage{times}
\usepackage{latexsym}

\usepackage[T1]{fontenc}

\usepackage[utf8]{inputenc}

\usepackage{microtype}

\usepackage{inconsolata}

\usepackage{graphicx}
\usepackage{multirow}
\usepackage[table]{xcolor}
\usepackage{algorithm}
\usepackage{algorithmic}
\usepackage{graphicx}
\usepackage{subfig}
\usepackage{amsmath}
\usepackage{amssymb}
\usepackage{booktabs}
\usepackage{tcolorbox}
\usepackage{listings}

\title{D2-ScaleAgent: Dual-Dimensional Scaling for 
Long Document Understanding}

\author{
    Hao Zhang$^1$, Longrong Yang$^2$, Lunhao Duan$^2$, Ziyang Wang$^3$, Qing-Guo Chen$^2$, Shanshan Zhao$^{2}$ \\
    $^1$Zhejiang University \\
    $^2$Alibaba Group \quad
    $^3$University of Science and Technology of China \\ 
}

\begin{document}
\maketitle
\begin{abstract}
Multi-modal retrieval-augmented generation (RAG) is a key technique for visually rich long document understanding. Existing multi-modal RAG methods are progressively advancing toward multi-agent systems: they first retrieve relevant pages based on a query, and then iteratively understand information within those pages. However, these methods typically rely on fixed workflows and lack the ability to dynamically scale computation at test time, often leading to insufficient evidence. 
To address this, we propose D2-ScaleAgent, an agentic framework that introduces a dual-dimensional scaling paradigm for retrieval and reasoning.
The core of D2-ScaleAgent is a Verifier agent-driven dynamic routing loop based on the intrinsic difficulty of the query, centered around a continuously updated evidence bank that serves as the agent's dynamic working memory: when retrieval needs to be expanded, the agent routes outward (retrieval scaling), decomposing the query into attributes and performing parallel page retrieval, followed by adaptive pruning to ensure comprehensive evidence coverage. When fine-grained reasoning is required, the agent routes inward (reasoning scaling), dynamically selecting sub-agents with varying granularity and count to extract evidence from pages. Finally, D2-ScaleAgent achieves logical closure over the evidence chain. Extensive experiments demonstrate that D2-ScaleAgent is effective on long and visually rich document benchmarks like MMLongBench-Doc, LongDocURL, etc.
\end{abstract}

\section{Introduction}
\begin{figure}[!t]
    \centering
    \includegraphics[width=0.95\linewidth]{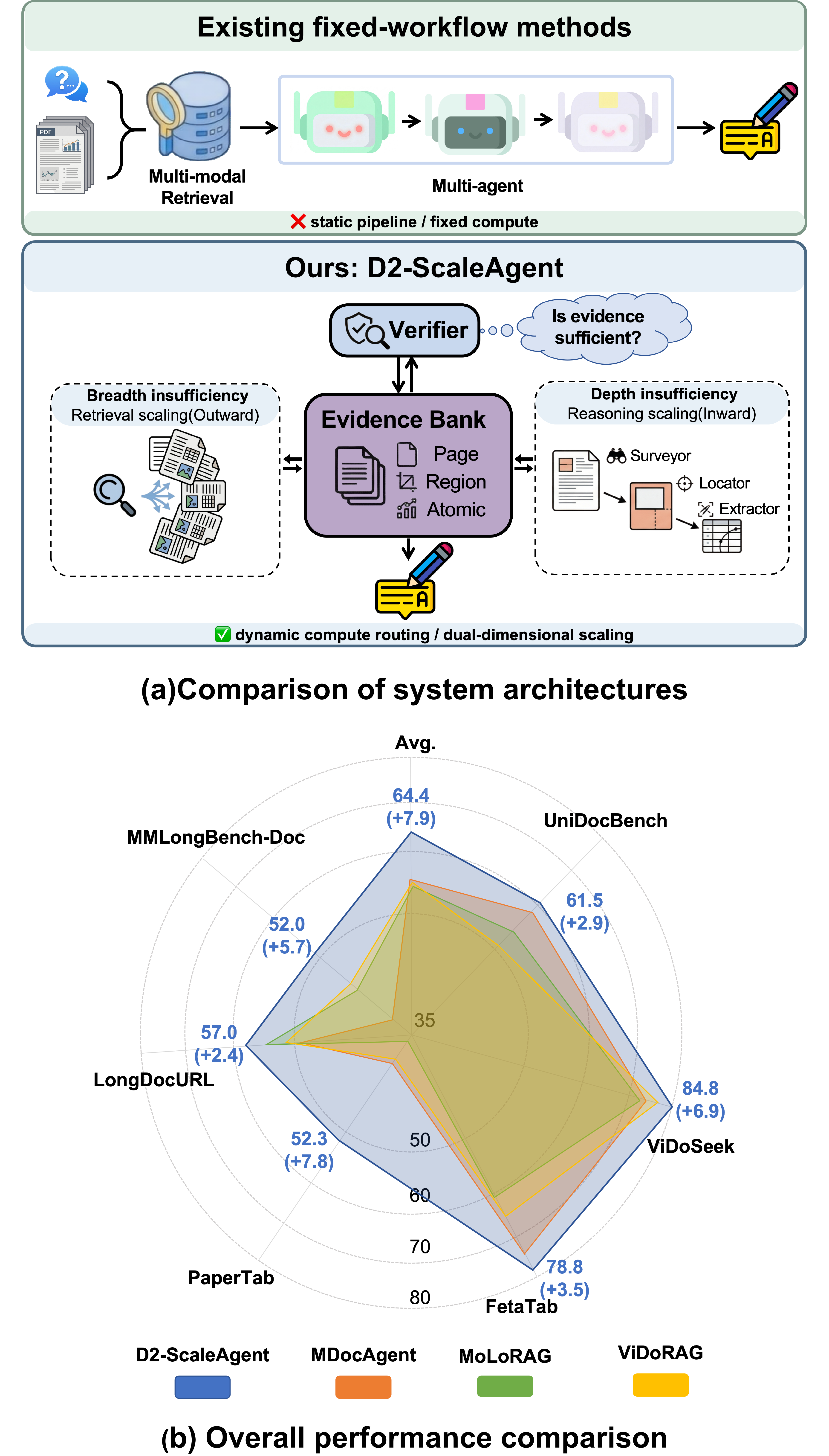}
    \caption{Architecture and performance of our proposed method, D2-ScaleAgent. (a) The framework achieves dual-dimensional scaling through a Verifier agent-driven dynamic routing loop, addressing evidence insufficiency by dynamically scaling outward for retrieval and inward for fine-grained understanding. (b) It shows improvements in comprehensive end-to-end QA accuracy across multiple long-document benchmarks, consistently outperforming existing baselines.}
    \label{figteaser}
     \vspace*{-12pt}
\end{figure}

Answering questions from visually rich long documents, like financial reports and scientific papers, is critical for many real-world applications~\cite{visdomrag,documentunderst,documentunderst2,Scientificrpapers,vdoc}. 
While Large Vision-Language Models (LVLMs) have made significant progress in general visual understanding \cite{li2023blip,liu2024llavanext,bai2023qwenvl, gemini, zhu2023minigpt}, they still struggle with visually rich long documents. A core conflict exists between the need for fine-grained visual details and the limited context window of LVLMs. This mismatch often leads to performance drops when processing long documents \cite{longvison,layoutlmv2,mmlongbench-doc}. Crucially, the main challenge of long-document understanding is not just the document's length, but the difficulty of finding "sufficient and adequate" evidence within limited computational resources \cite{vidorag, chen2025inter,multi-hopQA,metaembedTTS}.

To address this challenge, current research typically employs two primary strategies. The first involves multi-modal retrieval-augmented generation (RAG), which filters document content to first identify relevant pages before performing detailed reading \cite{vdocrag, ragreview}. The second route focuses on multi-agent or iterative mechanisms, which utilize multi-step analysis and reasoning to navigate complex document structures \cite{vidorag, agentsurvey}.
However, as illustrated in Figure \ref{figteaser}(a), these methods are often constrained by fixed workflows, acting more like fixed workflows than autonomous agents, lacking the flexibility to dynamically scale computation at test time based on the inherent difficulty of the user's query \cite{AdaptiveTTS}.
Consequently, many errors arise not because the model's reasoning template is flawed, but because the system fails to acquire sufficient evidence during generation \cite{ragerror}. The fundamental essence of long-document understanding failure is, therefore, an "Evidence Insufficiency Problem".

When human readers encounter insufficient information in a long document, they do not jump to conclusions \cite{pirolli1995information,kobayashi2024neural}. Instead, they distinguish between a failure to locate the correct page \cite{van2025wanting} and a failure to thoroughly process the relevant section \cite{li2024llatrievalverifiable, qian2026rvr}. Inspired by this, we categorize evidence insufficiency into two dimensions: (1) Breadth insufficiency, where incomplete coverage requires the system to scale outward to broader document contexts; and (2) Depth insufficiency, where localized but superficial understanding requires the system to scale inward for finer granularity and deeper reasoning. Consequently, rather than following a fixed path, long document understanding should dynamically scale across both dimensions based on detected evidence gaps. To this end, we propose D2-ScaleAgent, a dual-dimensional evidence-driven agentic framework.

D2-ScaleAgent abandons the typical static "retrieve-then-read" workflow, establishing instead a Verifier agent-driven closed-loop agent system centered around a continuously updated Evidence Bank, which acts as the agent's dynamic working memory. 
This Evidence Bank maintains the currently acquired page, region, and atomic evidence, recording their logical states: whether they support the answer, conflict, or remain missing. 
The core of this closed loop is a Verifier agent-driven dynamic routing. 
As shown in Figure~\ref{figteaser}(a), the Verifier serves as a trigger for the entire system, which strictly checks if the current Evidence Bank is sufficient to support a complete response. 
When an evidence gap is detected, the Verifier outputs a gap signal and dynamically routes the compute:
When new source pages are missing, it autonomously routes outward to retrieval scaling.
The retrieval scaling triggers an attribute decomposition of the original query, performing parallel candidate retrieval from multiple perspectives while applying rank fusion and adaptive pruning to strictly control context budgets. 
When fine-grained details are missing, it routes inward to reasoning scaling, dynamically selecting sub-agents of varying costs and granularities, including Global Surveyor, Region Locator, and Fine-grained Extractor.
Finally, the agentic system iteratively completes the evidence chain under the guidance of the Verifier until logical closure is achieved: a state where all identified evidence gaps are resolved to substantiate the final answer.

In summary, our major contributions are as follows:
\begin{itemize}
\item 
We identify the fundamental cause of failure in long document understanding as an "evidence insufficiency problem" and introduce the "Dual-Dimensional Scaling" paradigm. 
\item 
We propose the D2-ScaleAgent framework. Its primary contribution is a dynamic closed-loop mechanism, driven by a Verifier agent and centralized around an Evidence Bank that serves as the system's global memory, enabling profound synergy and seamless fallback between retrieval and reasoning operations.
\item 
Extensive evaluations on multiple complex, visually-rich long document benchmarks (e.g., MMLongBench-Doc~\cite{mmlongbench-doc}) demonstrate that D2-ScaleAgent achieves state-of-the-art performance, outperforming traditional RAG and static-agent workflows, as shown in Figure~\ref{figteaser}(b).
\end{itemize} 
\section{Related Work}

\textbf{Multi-modal Retrieval-Augmented Generation.}
Multi-modal RAG enhances document question-answering capabilities by supplying models with externally retrieved visual or textual contexts \cite{lewis2020retrieval,guu2020retrieval,borgeaud2022improving}. Recent studies have achieved significant advances in retrieval mechanisms. Representative examples include M3DocRAG \cite{m3docrag}, which visualizes PDF pages for end-to-end visual retrieval; MHier-RAG \cite{mhierrag}, which enables fine-grained localization within long documents via intra- and inter-page structural indices; and MoLoRAG \cite{molorag} and MegaRAG \cite{megarag}, which utilize page graphs or multi-modal knowledge graphs to model cross-page logical and entity relationships. Furthermore, ViDoRAG \cite{vidorag} attempts to move beyond fixed Top-K constraints through dynamic retrieval budget allocation, while VisDoMRAG \cite{visdomrag} models textual and visual evidence in parallel with consistency-constrained fusion. 
These works propel multi-modal document understanding toward joint visual-textual-structural modeling. However, existing multi-modal RAG methods typically rely on static retrieval and fixed dual-channel workflows. This lack of dynamic routing causes \textbf{Breadth Insufficiency}: failing to recall relevant information located elsewhere. Our framework addresses this via a \textit{Retrieval Scaling} module. Instead of blind Top-K expansion \cite{topk}, it decomposes the query into attributes and parallelizes retrieval from multiple perspectives, directing outward expansion specifically around missing evidence.

\textbf{Multi-Agent Systems.} Multi-agent systems have shown immense potential in handling reasoning over complex long documents \cite{zhao2024longagent,yao2022react,zhang2024chain}. These systems typically utilize specialized sub-agents with distinct divisions of labor to iteratively understand document content. For instance, MDocAgent \cite{mdocagent} enhances the mining of text-image details through multi-role collaboration, while DocAgent \cite{docagent} improves long-document processing capabilities via outline navigation, interactive reading, reviewer agents, and memory banks. MACT \cite{mact} introduces procedural test-time scaling, and ViDoRAG \cite{vidorag} adopts an iterative seeker-inspector-answer collaborative workflow. 
While multi-agent systems mitigate single-model limitations in complex DocQA, their predetermined workflows and fixed budgets struggle to dynamically allocate reasoning depth for fine-grained visual details. This rigidity causes \textbf{Depth Insufficiency}: incomplete evidence resolution. Our framework addresses this via a \textit{Verifier agent-driven dynamic routing loop} centered around an \textit{Evidence Bank}.
\begin{figure*}[htbp]
  \centering 
  \includegraphics[width=0.9\linewidth]{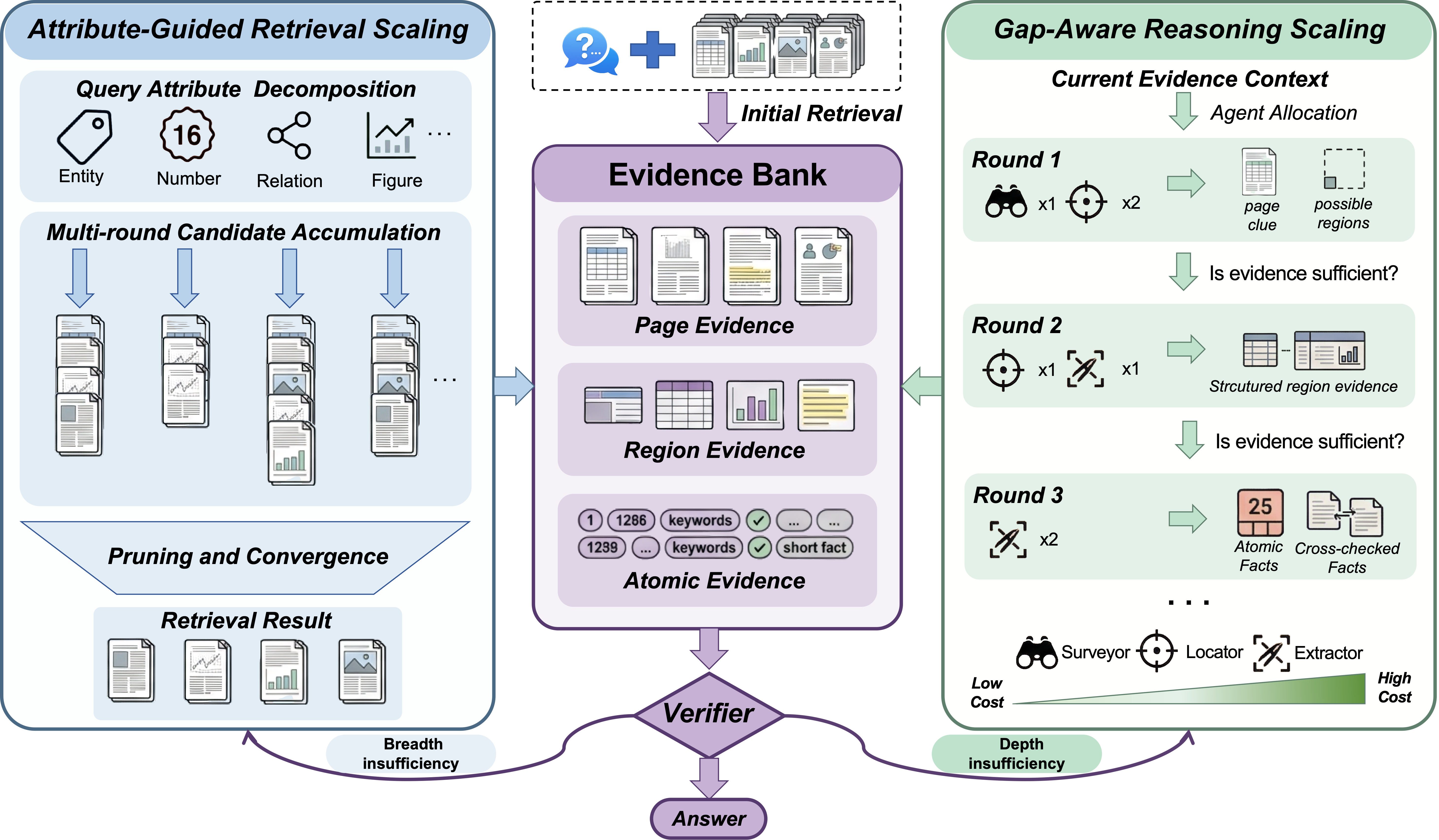} 
    \caption{Overview of D2-ScaleAgent. (1) The system first retrieves a candidate page pool and initializes the Evidence Bank. (2) A Verifier agent-driven loop then routes computation outward for retrieval scaling or inward for reasoning scaling based on the current evidence gap. (3) Multi-granularity evidence is continuously accumulated into the Evidence Bank. (4) Once logical closure is achieved, the system generates the final answer from the saturated evidence state.}
  \label{fig_overview}
\end{figure*}

\section{Methodology}
In this section, we present D2-ScaleAgent, a Dual-Dimensional Scaling framework that reformulates long-document understanding from a workflow into an on-demand compute routing problem, as illustrated in Figure \ref{fig_overview}.

\subsection{Problem Formulation}
We formulate long document understanding as an evidence-driven exploration. The system maintains a global \textbf{Evidence Bank ($\mathcal{B}_{t}$)} serving as the unified epistemic state at global system step $t$:
\begin{equation}
\mathcal{B}_{t}=\{E_t^{page}, E_t^{region}, E_t^{atomic}, s_{t}^{comp}, g_{t}\}
\end{equation}
These five variables respectively track the globally accumulated page-level, region-level, and atomic-level evidence, quantify the current evidence completeness score ($s_{t}^{comp}$), and explicitly record the current evidence gap ($g_{t}$). Guided by this unified state, the system dynamically allocates computational budgets to resolve the two primary failure modes in long-document reasoning: first introducing the attribute-guided retrieval scaling to resolve breadth insufficiency (Section 3.2), and subsequently presenting the gap-aware reasoning scaling to resolve depth insufficiency (Section 3.3).

\subsection{Attribute-Guided Retrieval Scaling}
Traditional static Top-$K$ retrieval paradigms are ill-equipped for complex visually-rich long documents, as the scope of required evidence varies drastically across queries. To address this, we formalize retrieval scaling as a dynamic, three-stage process aimed at adaptively expanding the search boundary until it autonomously converges to a high-value evidence set. Given a current query intent $q$ and a page space $\mathcal{P}$, the process operates as follows:

\textbf{Query Attribute Decomposition.} In visually-rich contexts, an information need is rarely singular. Instead of treating $q$ as a monolithic input, we leverage an LLM (whose prompt template is detailed in Appendix~\ref{prompts}) to decompose it into a weighted set of multi-perspective attribute queries:
\begin{equation}
\mathcal{Q} = \{(q_0, w_0), (q_1, w_1), \dots, (q_M, w_M)\}
\end{equation}
where $q_0$ is the core intent, $q_m$ are derived attribute perspectives, and $w_m$ represents the confidence weight ($\sum_{m=0}^{M} w_m = 1$).

\textbf{Multi-round Candidate Accumulation.} 
To gather complementary evidence, the system executes retrieval for the decomposed queries in $\mathcal{Q}$. At any internal retrieval step $j$ (where $0 \le j \le M$), the system processes the attribute query $q_j$ to retrieve a local candidate page set $\mathcal{R}^{(j)}$. It then updates a monotonically expanding global candidate page pool, defined as $\mathcal{C}^{(j)} = \bigcup_{m=0}^{j} \mathcal{R}^{(m)}$, which aggregates results from all queries processed up to step $j$. Since raw retrieval scores across different attribute perspectives are incomparable, we calculate a rank-based weighted fusion score for each accumulated candidate page $c$:

\begin{equation}
S_j(c) = \sum_{m=0}^{j} \mathbf{1}[c \in \mathcal{R}^{(m)}] \cdot \frac{w_m}{\kappa + r^{(m)}(c)}
\end{equation}
where $r^{(m)}(c)$ is the local rank of page $c$ in the $m$-th query's results, $w_m$ is the query's confidence weight, and $\kappa$ is a smoothing constant. This mechanism prioritizes high-value evidence consistently supported across multiple perspectives.

\textbf{Adaptive Pruning and Convergence.} To balance the expanding boundary with downstream reasoning costs, we introduce a confidence-aware pruning mechanism. We define the maximum accumulated score $S_j^{\max}$ and extract the high-value evidence set for the current round using a relative threshold $\alpha \in (0,1)$:
\begin{equation}
\mathcal{E}_j^{\mathrm{ret}} = \{c \in \mathcal{C}^{(j)} \mid S_j(c) \geq \alpha S_j^{\max}\}
\end{equation}
To autonomously determine if the retrieval scaling has saturated, we quantify the cross-round stability of this evidence set:
\begin{equation}
\text{Stab}_j = \frac{|\mathcal{E}_j^{\mathrm{ret}} \cap \mathcal{E}_{j-1}^{\mathrm{ret}}|}{|\mathcal{E}_j^{\mathrm{ret}}|}
\end{equation}
When $\text{Stab}_j$ consistently exceeds a predefined threshold $\tau$, the system infers that the retrieval frontier has converged. It dynamically terminates the internal expansion and designates this set as the global retrieval output for the global system step $t$: $\mathcal{E}_t^{\mathrm{ret}} \leftarrow \mathcal{E}_j^{\mathrm{ret}}$. 
In summary, the terminus of retrieval scaling is neither a fixed number of rounds nor a rigid Top-$K$ budget, but an adaptive stopping state achieved when the evidence set stabilizes under multi-perspective scrutiny.

\subsection{Gap-Aware Reasoning Scaling}
Even when relevant pages are successfully retrieved ($\mathcal{E}_t^{\mathrm{ret}}$), LVLMs frequently default to coarse semantic matching, struggling to parse precise numerical values or complex layouts from visually-rich contexts. To address this depth insufficiency, we introduce a Reasoning-Side Scaling mechanism. Rather than indiscriminately forcing pages through a fixed workflow, this mechanism dynamically invokes a toolkit of cost-stratified cognitive sub-agents based on explicit evidence gaps, executing a coarse-to-fine transition from macroscopic intuition to auditable atomic facts. The sub-agents' prompt templates are detailed in Appendix~\ref{prompts}. The reasoning toolkit comprises three hierarchical levels of sub-agents, which are adaptively selected by the downstream routing mechanism:

\textbf{Global Surveyor: Low-cost Page Understanding.} The Surveyor executes a macroscopic scan over the candidate set $\mathcal{E}_t^{\mathrm{ret}}$ to establish a directional semantic prior, preventing the premature allocation of expensive compute to irrelevant details. It outputs incremental page-level evidence $e_t^{\mathrm{page}}$:
\begin{equation}
e_t^{\mathrm{page}} = f_{\mathrm{sur}}(q, \mathcal{E}_t^{\mathrm{ret}})
\end{equation}

\textbf{Region Locator: Medium-cost Region Focusing.} Acting as a dimensionality reduction mechanism, the Locator isolates critical regions from the broad candidate set and generates structured summaries (e.g., identifying specific tables), outputting a highly relevant page subset $\mathcal{E}_t^{\mathrm{key}}$ and region-level evidence $e_t^{\mathrm{region}}$:
\begin{equation}
(\mathcal{E}_t^{\mathrm{key}}, e_t^{\mathrm{region}}) = f_{\mathrm{loc}}(q, \mathcal{E}_t^{\mathrm{ret}})
\end{equation}

\textbf{Fine-grained Extractor: High-cost Atomic Extraction.} For the identified critical regions, this module undertakes high-resolution precision reading based on an explicit extraction specification ($\psi_t$) dynamically generated from current evidence gaps. It extracts specific, verifiable atomic facts $e_t^{\mathrm{atomic}}$:
\begin{equation}
e_t^{\mathrm{atomic}} = f_{\mathrm{ext}}(q, \mathcal{E}_t^{\mathrm{ret}}, \psi_t)
\end{equation}

Crucially, the multi-granularity evidence extracted by these adaptive operations is not held in isolated caches. The incremental evidence ($e_t^{*}$) is continuously appended to the globally accumulated sets ($E_t^{*}$):
\begin{equation}
E_{t+1}^{*} = E_t^{*} \cup \{e_t^{*}\}
\end{equation}
Consequently, the global epistemic state of the Evidence Bank is explicitly and incrementally updated at each reasoning step:
\begin{equation}
\mathcal{B}_{t+1} = \mathrm{Update}(\mathcal{B}_t, e_t^{\mathrm{page}}, e_t^{\mathrm{region}}, e_t^{\mathrm{atomic}})
\end{equation}
This evidence-driven update mechanism formalizes the epistemic state, providing the global Verifier with a unified basis to drive subsequent inward/outward routing decisions.

\subsection{Verifier Agent-Driven Routing}
The orchestration of this evidence-driven computation is governed by the Verifier, which functions not merely as a posterior checker, but as the system's explicit stopping and expansion trigger. The Verifier's prompt template is detailed in Appendix~\ref{prompts}. At each inference step $t$, the Verifier assesses the epistemic completeness of the Evidence Bank, identifying specific logical discontinuities or cross-modal contradictions:
\begin{equation}
(s_t^{comp}, g_t) = f_{ver}(q, \mathcal{B}_t)
\end{equation}
These evaluations are subsequently written back to update the global state: $\mathcal{B}_t \leftarrow \text{Update}_{ver}(\mathcal{B}_t, s_t^{comp}, g_t)$. Based on this explicit gap $g_t$, the system triggers one of two distinct routing dimensions:

\textbf{1. Inward Digging (Addressing Depth Insufficiency).} When the critical source pages are located but parsed localized facts remain insufficient, the routing mechanism dynamically allocates a specific cognitive operation $o_t$ from the reasoning toolkit (Section 3.3) to extract deeper incremental evidence:
\begin{equation}
\begin{split}
\text{if } g_t \in \mathcal{G}_{depth}, \quad & o_t \leftarrow \text{Reasoning-Scale}(g_t, \mathcal{B}_t), \\
& o_t \in \{f_{sur}, f_{loc}, f_{ext}\}
\end{split}
\end{equation}

\textbf{2. Outward Expansion (Addressing Breadth Insufficiency).} If the Verifier reveals an unbridgeable contextual void indicating missing foundational source evidence, the system translates the explicit gap $g_t$ into a novel retrieval intent to dynamically re-expand the candidate boundary via the retrieval toolkit (Section 3.2):
\begin{equation}
\begin{split}
\text{if } g_t \in \mathcal{G}_{breadth}, \quad & q_{t}^{new} \leftarrow \Phi(g_t, q), \\
& \mathcal{E}_{t}^{ret} \leftarrow \text{Retrieval-Scale}(q_{t}^{new})
\end{split}
\end{equation}

The closed-loop routing mechanism operates continuously until a rigorous termination condition is met:
\begin{equation}
\text{Stop if } s_t^{comp} \geq \delta \text{ and } g_t = \varnothing
\end{equation}
where $\delta$ defines the strict evidence closure threshold. Once this condition is satisfied, the system decisively breaks the iterative loop and generates the final answer based on the logically saturated Evidence Bank:
\begin{equation}
a = f_{ans}(q, \mathcal{B}_t)
\end{equation}


By conditioning generation strictly on a verified, logically closed evidence state rather than forcing an output under incomplete information, the system avoids hallucinating unsupported conclusions. Further details are provided in Appendix~\ref{Algorithmic}.

\section{Experiments}

\subsection{Experiment Setup}

\textbf{Datasets}: To cover a wide range of real-world application scenarios, we selected six multi-modal long-document benchmarks: MMLongBench-Doc \cite{mmlongbench-doc}, LongDocURL \cite{longdocurl}, PaperTab \cite{udabench}, FetaTab \cite{udabench}, ViDoSeek \cite{vidorag} and UniDoc-Bench \cite{unidoc}. These datasets span both open and domain-specific fields, encompassing varying text lengths and rich visual elements to comprehensively test models' multi-modal understanding capabilities in long contexts.

\begin{table*}[t]
  \centering
    \small 
  \renewcommand{\arraystretch}{0.9} 
  \setlength{\tabcolsep}{3 pt} 
  \caption{Performance comparison of different methods across multiple benchmarks. VQA takes the entire document as image input.}
  \label{tab:main_results}
  \begin{tabular}{lccccccc}
    \toprule
    Method & MMLongBench-Doc & LongDocURL & PaperTab & FetaTab & ViDoSeek & UniDoc-Bench &  Avg \\
    \midrule
    \multicolumn{8}{c}{GPT-4o} \\
    \midrule 
    VQA & 43.7 & 51.1 & 31.3 & 75.7 & 78.8 & 56.1 & 56.1 \\
    ViDoRAG \cite{vidorag} & 44.4 & 50.8 & 30.6 & 66.1 & 70.7 & 58.0 & 53.4 \\
    MoLoRAG \cite{molorag} & 40.7 & 55.5 & 29.3 & 64.7 & 74.1 & 57.1 & 53.6 \\
    MDocAgent \cite{mdocagent} & 34.7 & 50.9 & 45.5 & 79.8 & 77.4 & 61.2 & 58.3 \\
    D2-ScaleAgent & \textbf{52.0} & \textbf{56.0} & \textbf{47.5} & \textbf{82.8} & \textbf{81.8} & \textbf{62.2} & \textbf{63.7} \\

    \midrule
    \multicolumn{8}{c}{Gemini-3-flash-preview} \\
    \midrule 
    VQA & 65.0 & 65.4 & 58.6 & 72.7 & 86.8 & 60.6 & 68.2 \\
    ViDoRAG \cite{vidorag} & \textbf{65.3} & 60.6 & 56.5 & 80.8 & 83.8 & 62.9 & 68.3 \\
    MoLoRAG \cite{molorag} & 49.2 & 59.1 & 48.5 & 67.7 & 76.3 & 66.3 & 61.2 \\
    MDocAgent \cite{mdocagent} & 42.6 & 45.0 & 54.5 & 80.8 & 76.3 & 64.9 & 60.7 \\
    D2-ScaleAgent & 63.0 & \textbf{64.4} & \textbf{62.6} & \textbf{84.8} & \textbf{85.8} & \textbf{73.5} & \textbf{72.4} \\

    \midrule
    \multicolumn{8}{c}{Qwen2.5-VL-7B-Instruct} \\
    \midrule 
    VQA & 32.9 & 44.6 & 34.3 & 61.6 & 70.7 & 35.7 & 46.6 \\
    ViDoRAG \cite{vidorag} & 30.5 & 43.5 & 33.0 & 48.0 & 79.5 & 38.9 & 45.6 \\
    MoLoRAG \cite{molorag} & 38.1 & 48.6 & 30.3 & 55.6 & 69.9 & 46.9 & 48.2 \\
    MDocAgent \cite{mdocagent} & 32.2 & 43.9 & 32.3 & 66.7 & 68.8 & \textbf{51.0} & 49.2 \\
    D2-ScaleAgent & \textbf{42.1} & \textbf{49.4} & \textbf{43.4} & \textbf{70.7} & \textbf{81.8} & 47.0 & \textbf{55.7} \\

    \midrule
    \multicolumn{8}{c}{Qwen3-VL-8B-Instruct} \\
    \midrule 
    VQA & 49.0 & 56.8 & 49.5 & 69.0 & 82.8 & 53.0 & 60.0 \\
    ViDoRAG \cite{vidorag} & 44.7 & 53.0 & 54.6 & 68.4 & 77.6 & 51.0 & 58.2 \\
    MoLoRAG \cite{molorag} & 46.0 & 55.0 & 36.4 & 57.6 & 72.0 & 51.0 & 53.0 \\
    MDocAgent \cite{mdocagent} & 40.7 & 52.2 & 45.5 & 73.7 & 76.3 & 57.1 & 57.6 \\
    D2-ScaleAgent & \textbf{51.1} & \textbf{58.1} & \textbf{55.5} & \textbf{76.7} & \textbf{89.9} & \textbf{63.2} & \textbf{65.8} \\
    \bottomrule
  \end{tabular}
\end{table*}


\textbf{Baselines}: D2-ScaleAgent is benchmarked against 3 advanced open-source baselines categorized into two mainstream paradigms: (1) \textbf{Multi-modal RAG}:  MoLoRAG \cite{molorag}. (2) \textbf{Multi-agent Systems}: This includes MDocAgent \cite{mdocagent} and ViDoRAG \cite{vidorag}, representing advanced frameworks that operate on fixed workflows.

More implementation details are provided in Appendix~\ref{Implementation_details}.


\subsection{Main Results}

\textbf{(1) End-to-End QA Performance.}
Table \ref{tab:main_results} reports the end-to-end QA accuracy of D2-ScaleAgent and all baseline methods on the six benchmarks. Overall, D2-ScaleAgent achieves the best average performance among all methods, consistently outperforming both multi-modal RAG and fixed-workflow multi-agent baselines across all evaluated models (e.g., 63.7 vs. MDocAgent's 58.3 on GPT-4o). Notably, VQA, which processes the full document image directly without relying on explicit retrieval or iterative reasoning stages, outperforms several retrieval-based baselines in multiple settings, suggesting that stronger foundation models can substantially narrow the gap between holistic VQA and fixed retrieval-and-reasoning pipelines. This trend is particularly evident with Gemini-3-flash-preview: VQA even outperforms D2-ScaleAgent on MMLongBench-Doc \cite{mmlongbench-doc} and LongDocURL \cite{longdocurl}. We emphasize, however, that the advantage of D2-ScaleAgent does not lie in uniformly outperforming VQA in every setting, but in mitigating the brittleness of fixed retrieval and reasoning workflows under evidence-insufficient scenarios. These gains are particularly prominent on benchmarks with complex layouts and dispersed evidence.

\begin{figure}[htbp]
  \centering 
  \includegraphics[width=\linewidth]{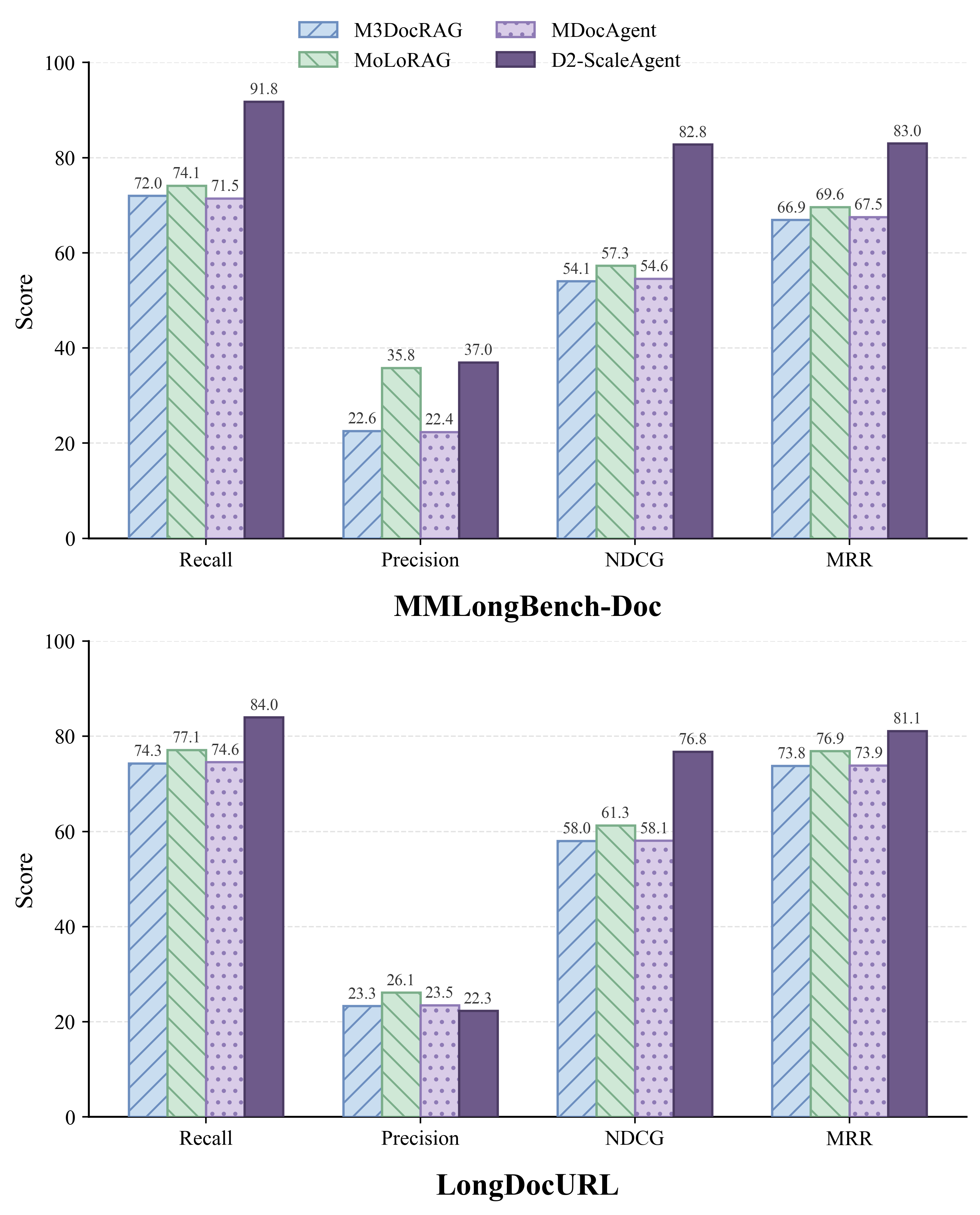} 
  \caption{Retrieval performance comparison on MMLongBench-Doc and LongDocURL datasets.}
  \label{fig_retrieval}
\end{figure}

\textbf{(2) Retrieval Performance.}
We further evaluate retrieval quality on MMLongBench-Doc \cite{mmlongbench-doc} and LongDocURL \cite{longdocurl}. As shown in Figure \ref{fig_retrieval}, the baseline results are taken from MoLoRAG \cite{molorag}, and D2-ScaleAgent consistently improves Recall, Precision, nDCG, and MRR over strong baselines. These gains indicate that our method not only retrieves more relevant evidence, but also ranks it more effectively. The improvements mainly come from the retrieval-side scaling strategy, which performs query attribute decomposition and iterative evidence expansion rather than relying on a single static Top-K retrieval step. 
This allows the system to better handle questions whose supporting evidence is distributed across multiple pages or only partially captured by the initial query. 
Overall, the results confirm the importance of coupling retrieval expansion with downstream reasoning.

\begin{table}[htbp]
  \centering
  \caption{Ablation study of different components on the MMLongBench-Doc dataset.}
  \label{tab:ablation_study}
  \resizebox{\linewidth}{!}{
    \begin{tabular}{cccccc}
      \toprule
      \multirow{2}{*}{\shortstack{Retrieval Scaling}} & \multicolumn{3}{c}{Reasoning Scaling} & \multirow{2}{*}{\shortstack{Verifier-\\Driven Loop}} & \multirow{2}{*}{MMLongBench-Doc} \\
      \cmidrule(lr){2-4}
      & Surveyor & Locator & Extractor & & \\
      \midrule
      Evidence & $\checkmark$ & $\checkmark$ & $\checkmark$ & $\checkmark$ & 54.9 \\
      \midrule
      $\checkmark$ & $\checkmark$ & $\checkmark$ & $\checkmark$ & $\checkmark$ & \textbf{52.0} \\
      $\boldsymbol{\times}$ & $\checkmark$ & $\checkmark$ & $\checkmark$ & $\checkmark$ & 46.8 \\
      $\checkmark$ & $\boldsymbol{\times}$ & $\checkmark$ & $\checkmark$ & $\checkmark$ & 46.5 \\
      $\checkmark$ & $\checkmark$ & $\boldsymbol{\times}$ & $\checkmark$ & $\checkmark$ & 47.1 \\
      $\checkmark$ & $\checkmark$ & $\checkmark$ & $\boldsymbol{\times}$ & $\checkmark$ & 47.5 \\
      $\checkmark$ & $\checkmark$ & $\checkmark$ & $\checkmark$ & $\boldsymbol{\times}$ & 44.1 \\
      \midrule
      $\checkmark$ & \multicolumn{4}{c}{GPT-4o Answer Directly} & 45.0 \\
      \bottomrule
    \end{tabular}
  } 
\end{table}

\begin{figure*}[htbp]
  \centering 
  \includegraphics[width=\linewidth]{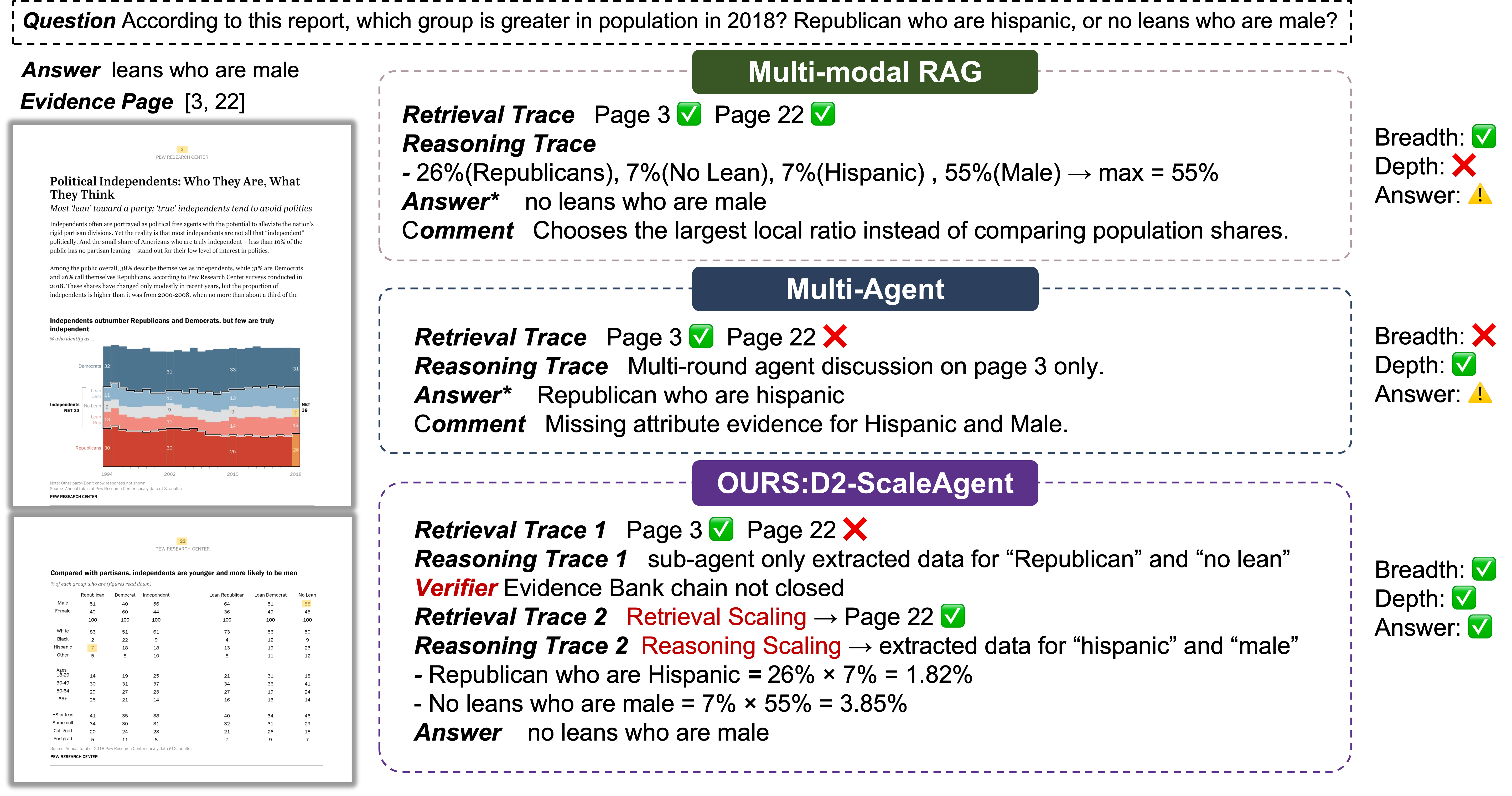} 
    \caption{Case study of D2-ScaleAgent. Given a cross-page compositional question comparing population sizes, both baseline methods fail due to either breadth or depth insufficiency. In contrast, D2-ScaleAgent, guided by the Verifier agent-driven dynamic routing loop, identifies that the initial evidence chain is incomplete, then triggers Attribute-Guided Retrieval Scaling to locate the missing page and Gap-Aware Reasoning Scaling to extract the required fine-grained numerical evidence. }
  \label{Case_Study}
\end{figure*}

\subsection{Ablation Studies}
As shown in Table \ref{tab:ablation_study}, we ablate D2-ScaleAgent on MMLongBench-Doc to quantify each component's contribution. The full model (52.0) outperforms the naive GPT-4o baseline (45.0) and approaches the upper bound of 54.9, which is achieved by replacing the retrieval module with ground-truth evidence. Crucially, disabling the Verifier agent-driven loop causes the steepest accuracy drop (to 44.1), underscoring the critical role of rigorous evidence verification and closed-loop routing. Similarly, discarding the Retrieval-Side scaling (46.8) or removing any of the inward-scaling reasoning agents—Surveyor (46.5), Locator (47.1), and Extractor (47.5)—consistently degrades performance. 
These findings demonstrate that static workflows are insufficient; dynamic retrieval expansion, multi-granularity visual parsing, and rigorous evidence verification must be coordinated to resolve complex long document understanding tasks. Additional ablation results are provided in Appendix~\ref{ablation_extended}.

\subsection{Compute-cost analysis}
\label{Compute-cost analysis}
We have profiled D2-ScaleAgent on two representative benchmarks, MMLongBench-Doc and ViDoSeek. As shown in Table~\ref{tab:compute_cost}, computational costs dynamically scale with query difficulty: MMLongBench-Doc requires more verification and routing (21.4K tokens, 16.22s) due to its structural complexity and dispersed evidence, whereas ViDoSeek exhibits lower overhead (15.9K tokens, 11.89s) with fewer iterative corrections. D2-ScaleAgent adaptively allocates computation based on evidence insufficiency.
All reported efficiency metrics are averaged per query, and latency is measured end-to-end using the GPT-4o API.

\begin{table}[htbp]
  \centering
  \caption{Compute-cost analysis and profiling of D2-ScaleAgent on MMLongBench-Doc and ViDoSeek using GPT-4o.}
  \label{tab:compute_cost}
  \resizebox{\linewidth}{!}{
    \begin{tabular}{lcc}
      \toprule
      \textbf{Metric} & \textbf{MMLongBench-Doc} & \textbf{ViDoSeek} \\
      \midrule
      Accuracy & 52.0 & 81.8 \\
      Retrieval rounds & 1.33 & 1.12 \\
      Reasoning calls (S/L/E) & 1.51 / 1.12 / 0.63 & 1.75 / 0.34 / 0.47 \\
      Verifier calls & 1.76 & 1.14 \\
      Routing-agent calls & 5.02 & 3.70 \\
      Tokens & 21.4K & 15.9K \\
      E2E latency & 16.22 s & 11.89 s \\
      \bottomrule
    \end{tabular}
  }
\end{table}

\subsection{Case Study}
Figure \ref{Case_Study} presents an example of cross-page question answering. The question cannot be resolved from a single page, as it requires reasoning over the 2018 group distribution on Page 3 and the demographic composition table on Page 22. Multi-modal RAG successfully retrieves both pages, but fails to perform the required cross-page composition; instead, it compares local ratios and arrives at the correct answer only by chance. Fixed Multi-agent exhibits stronger local reasoning ability, yet remains confined to Page 3 and therefore misses the critical attribute evidence on Page 22. In contrast, D2-ScaleAgent identifies the incomplete evidence chain through the Verifier, triggers retrieval scaling to acquire the missing page, and applies fine-grained reasoning to complete the population-level comparison, thereby producing the correct answer.
This case highlights that D2-ScaleAgent does not merely benefit from stronger reasoning or broader retrieval in isolation, but from its Verifier agent-driven coordination of the two, which enables the system to adaptively close the evidence gap and achieve logical closure over a distributed evidence chain.

\section{Conclusions}
In this paper, we tackle the long document understanding task by addressing the limitations of prior methods that rely on fixed workflows and suffer from evidence insufficiency. By introducing a dual-dimensional scaling paradigm, our D2-ScaleAgent utilizes a Verifier agent-driven dynamic routing loop centered on a continuously updated evidence bank to perform both outward retrieval scaling and inward reasoning scaling. Extensive experiments demonstrate that D2-ScaleAgent achieves logical closure over evidence chains and delivers SOTA performance on complex benchmarks like MMLongBench-Doc, LongDocURL, etc.

\clearpage 
\newpage

\section*{Limitations}
While D2-ScaleAgent effectively mitigates evidence insufficiency through dual-dimensional scaling, this paradigm inherently trades computational efficiency for reasoning accuracy. A primary limitation of our framework is the inevitable increase in inference latency and computational overhead. The dynamic routing mechanism, which necessitates multi-round attribute-guided retrieval (outward scaling) and the on-demand invocation of cost-stratified cognitive sub-agents (inward scaling), results in execution times significantly longer than standard single-pass inference models. Consequently, the current architecture is computationally intensive and may be suboptimal for real-time applications or severely latency-constrained deployment scenarios. A comprehensive profiling of this computational efficiency trade-off is detailed in subsection~\ref{Compute-cost analysis}.

\section*{Ethical Considerations}
The datasets utilized in this work do not contain any personally identifiable or sensitive information, as all materials were sourced exclusively from publicly available domains. Furthermore, the curation, processing, and refinement of the data were conducted in strict adherence to applicable copyright laws and intellectual property guidelines.

\section*{The Use of AI assistants} AI assistants (ChatGPT) are used to correct potential grammatical inaccuracies in the manuscript. AI assistants do not participate in research ideation.


\bibliography{custom}

\clearpage 
\newpage 

\appendix
\section{Detailed Algorithmic Workflow}
\label{Algorithmic}
\begin{algorithm}
\caption{Process of \textbf{D2-ScaleAgent}}
\label{alg:d2_scaleagent}
\begin{algorithmic}[1]
\raggedright 
\STATE Initialize query $q$, document space $\mathcal{P}$, Evidence Bank $\mathcal{B}_0$;
\STATE Set step $t = 0$;
\STATE Get initial evidence $\mathcal{E}_0^{ret}$ via $\text{Retrieval-Scale}(q, \mathcal{P})$
\STATE Update $\mathcal{B}_0$ with initial evidence $\mathcal{E}_0^{ret}$

\WHILE{\textbf{true}}
    \STATE \textcolor{red}{\textit{\# Verifier assessment}}
    \STATE Assess completeness and gap: $(s_t^{comp}, g_t) \gets f_{ver}(q, \mathcal{B}_t)$
    
    \IF{$s_t^{comp} \geq \delta$ \textbf{and} $g_t = \varnothing$}
        \STATE \textbf{break}
    \ENDIF
    
    \STATE \textcolor{red}{\textit{\# Inward / Outward routing}}
    \IF{$g_t \in \mathcal{G}_{depth}$}
        \STATE Route inward: allocate reasoning operator $o_t \in \{f_{sur}, f_{loc}, f_{ext}\}$
        \STATE Extract incremental evidence $e_t$ via reasoning operation
    \ELSIF{$g_t \in \mathcal{G}_{breadth}$}
        \STATE Route outward: derive new query $q_t^{new} \gets \Phi(g_t, q)$
        \STATE Extract evidence $e_t$ via $\text{Retrieval-Scale}(q_t^{new})$
    \ENDIF
    
    \STATE \textcolor{red}{\textit{\# Evidence integration and state transition}}
    \STATE $\mathcal{B}_{t+1} \gets \text{Update}(\mathcal{B}_t, e_t)$
    \STATE $t \gets t + 1$
\ENDWHILE

\STATE \textbf{return} Final answer $a \gets f_{ans}(q, \mathcal{B}_t)$
\end{algorithmic}
\end{algorithm}

The inference process of D2-ScaleAgent is summarized in Algorithm \ref{alg:d2_scaleagent}. Unlike traditional methods that rely on fixed retrieval budgets (e.g., rigid Top-K selection), this algorithm operates through an adaptive, Verifier agent-driven closed-loop. By maintaining a global Evidence Bank ($\mathcal{B}_t$), the algorithm dynamically evaluates the current epistemic completeness ($s_t^{comp}$) and identifies specific evidence gaps ($g_t$). Leveraging a dual-dimensional routing mechanism, the traversal is precisely directed: it routes outward (retrieval scaling) to expand the search space when encountering breadth insufficiency, and routes inward (reasoning scaling) to extract fine-grained visual details when facing depth insufficiency. This on-demand allocation ensures scalability and avoids the computational waste of processing irrelevant pages. The iterative execution strictly terminates only when logical chain closure is achieved ($s_t^{comp} \ge \delta$ and $g_t = \varnothing$). The final output is a highly accurate answer, directly grounded in a rigorously verified and sufficient set of multi-granularity evidence.

\section{Case Study for Execution Traces}
\begin{figure*}
  \centering 
  \includegraphics[width=1.0\linewidth]{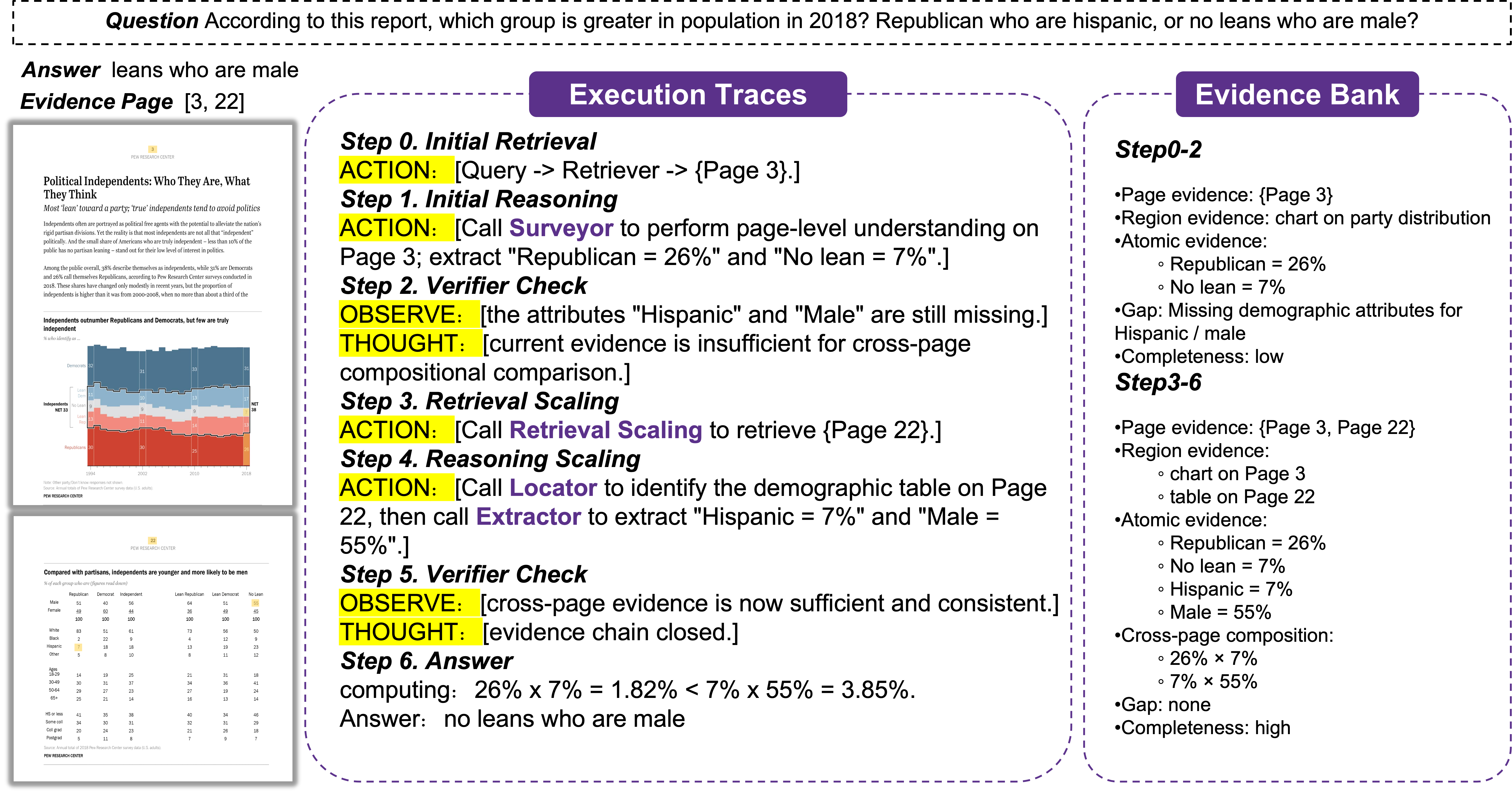}
    \caption{A case study illustrating the dual-dimensional scaling behavior of D2-ScaleAgent. Starting from an incomplete initial retrieval result, the agent identifies missing evidence through Verifier feedback and dynamically scales outward to retrieve additional pages, and inward to extract fine-grained attributes from them. The execution trace shows how the agent progressively gathers cross-page evidence from the evidence bank and finally derives the correct answer through compositional reasoning.}
  \label{fig_trace}
\end{figure*}

As shown in Figure \ref{fig_trace}, we illustrate the execution traces of D2-ScaleAgent with a cross-page compositional QA example. Given the query, the system first performs initial retrieval and identifies Page 3 as the starting evidence source. The Surveyor then extracts partial evidence related to party distribution, while the Verifier determines that the evidence chain is still incomplete because the key attributes “Hispanic” and “Male” are missing. Guided by this gap signal, D2-ScaleAgent activates Retrieval-side scaling to expand the search boundary and retrieve Page 22. It then invokes Reasoning-side scaling, where the Locator identifies the relevant demographic table and the Extractor obtains the missing atomic evidence required for cross-page composition. As the retrieved and extracted evidence is continuously written into the Evidence Bank, the Verifier re-evaluates the accumulated support until logical closure is achieved. The system then performs the final compositional comparison and produces the correct answer. This case highlights how D2-ScaleAgent dynamically coordinates retrieval expansion and fine-grained reasoning under Verifier guidance to resolve evidence insufficiency in long-document understanding.

\section{Extended Ablation Study on Diverse Benchmarks}
\label{ablation_extended}
To further validate the effectiveness of each component across different document types and evidence distributions, we extend our ablation study to two additional benchmarks: PaperTab (a table-centric benchmark) and UniDoc-Bench (a comprehensive multi-modal benchmark). Results are summarized in Table \ref{tab:ablation_extended}.

\begin{table}[htbp]
    \centering
    \caption{Extended ablation study of different components on PaperTab and UniDoc-Bench datasets.}
    \label{tab:ablation_extended}
    \resizebox{\linewidth}{!}{
      \begin{tabular}{ccccccc}
        \toprule
        \multirow{2}{*}{\shortstack{Retrieval\\Scaling}} & \multicolumn{3}{c}{Reasoning Scaling} & \multirow{2}{*}{\shortstack{Verifier-\\Driven Loop}} & \multirow{2}{*}{PaperTab} & \multirow{2}{*}{UniDoc-Bench} \\
        \cmidrule(lr){2-4}
        & Surveyor & Locator & Extractor & & & \\
        \midrule
        Evidence & $\checkmark$ & $\checkmark$ & $\checkmark$ & $\checkmark$ & X & 65.1 \\
        \midrule
        $\checkmark$ & $\checkmark$ & $\checkmark$ & $\checkmark$ & $\checkmark$ & \textbf{47.5} & \textbf{62.2} \\
        $\boldsymbol{\times}$ & $\checkmark$ & $\checkmark$ & $\checkmark$ & $\checkmark$ & 42.4 & 59.2 \\
        $\checkmark$ & $\boldsymbol{\times}$ & $\checkmark$ & $\checkmark$ & $\checkmark$ & 43.3 & 59.4 \\
        $\checkmark$ & $\checkmark$ & $\boldsymbol{\times}$ & $\checkmark$ & $\checkmark$ & 44.3 & 60.2 \\
        $\checkmark$ & $\checkmark$ & $\checkmark$ & $\boldsymbol{\times}$ & $\checkmark$ & 45.6 & 61.3 \\
        $\checkmark$ & $\checkmark$ & $\checkmark$ & $\checkmark$ & $\boldsymbol{\times}$ & 42.9 & 58.7 \\
        \midrule
        $\checkmark$ & \multicolumn{4}{c}{GPT-4o Answer Directly} & 39.1 & 57.4 \\
        \bottomrule
      \end{tabular}
    } 
  \end{table}

Note: The oracle-evidence result is unavailable for PaperTab because it does not provide ground-truth evidence-page annotations.




\section{Details}
\label{Implementation_details}

\subsection{Implementation Details}
For visual embedding, we utilize ColQwen2-v1.0 \cite{colpali}. The D2-ScaleAgent is designed with a plug-and-play architecture for foundation models; in our primary experiments, we default to GPT-4o \cite{gpt4o} as the core reasoning engine, which powers all functional modules including the Surveyor, Locator, Extractor, and Verifier. Additional foundational models tested include Gemini-3-flash-preview \cite{gemini}, Qwen2.5-VL-7B-Instruct \cite{qwen2.5VL}, and Qwen3-VL-8B-Instruct \cite{qwen3vl}. The entire agentic routing is implemented using the smolagents framework \cite{smolagents}. 

\subsection{Metrics Details}
We evaluate performance across generation and retrieval dimensions. For MMLongBench-Doc \cite{mmlongbench-doc} and LongDocURL \cite{longdocurl}, we employ GPT-4o \cite{gpt4o} to extract answers and evaluate using Exact Match (EM) and Accuracy. For PaperTab \cite{udabench}, FetaTab \cite{udabench}, and UniDoc-Bench \cite{unidoc}, we utilize GPT-4o for Binary Correctness (0/1) scoring. For ViDoSeek \cite{vidorag}, we utilize GPT-4o as the judge model to score the generated final answers against the ground truth reference on a scale of 1 to 5; an output achieving a score of 4 or higher is considered correct \cite{llmjudger}. To assess retrieval quality, we report Recall, Precision, Normalized Discounted Cumulative Gain (NDCG), and Mean Reciprocal Rank (MRR).

\subsection{Hyperparameters}
This section provides the default values of the hyperparameters used in D2-ScaleAgent throughout our experiments, as summarized in Table~\ref{tab:hyperparameters}.

\begin{table}[htbp]
  \centering
  \caption{Default values of hyperparameters used in D2-ScaleAgent.}
  \label{tab:hyperparameters}
  \begin{tabular}{lc}
    \toprule
    \textbf{Hyperparameter} & \textbf{Default Value} \\
    \midrule
    Pruning threshold ($\alpha$) & 0.7 \\
    Stability threshold ($\tau$) & 0.8 \\
    Closure threshold ($\delta$) & 8 \\
    Maximum decomposed queries & 6 \\
    Maximum retrieval rounds & 5 \\
    Maximum reasoning rounds & 20 \\
    Maximum sub-agent retries & 3 \\
    LLM token limit per call & 8,192 \\
    Top-$k$ & None (adaptive) \\
    \bottomrule
  \end{tabular}
\end{table}

\section{Prompt Templates for D2-ScaleAgent}
\label{prompts}
This section details the specific prompt templates designed to drive the dual-dimensional scaling process in D2-ScaleAgent. Below, we present the complete prompt for the Query Attribute Decomposition, Global Surveyor, Region Locator, Fine-grained Extractor, and the Verifier.

\begin{figure*}[htbp] 
    \centering
\noindent  The prompt for Query Attribute Decomposition is provided below:
\begin{tcolorbox}[     
    colback=white, 
    colframe=gray,               
    coltitle=white,                    
    colbacktitle=gray,                  
    title=\textbf{Prompt for Query Attribute Decomposition} 
]
   
    \textbf{\# Task Description  \# }
    
    You are an expert query decomposer for visually-rich long document retrieval. In visually-rich contexts, an information need is rarely singular. Your task is to decompose the user's question into a set of multi-perspective attribute queries.

    \vspace*{5pt}

    \textbf{\# Guidelines \#}
    
    \begin{itemize}
        \item \textbf{Dynamic Angles:} Autonomously define retrieval angles (e.g., entity, visual, comparative, numerical) based on the question to maximize recall across different document layouts.
        \item \textbf{Query Format:} Generate short, retrieval-friendly keyword phrases. Avoid pronouns. Include concrete anchors (entities, time ranges, structural hints).
        \item \textbf{Confidence Scores:} Assign a confidence score (\texttt{conf}) to each query. Scores must be floats with two decimal places and sum to exactly 1.00.
    \end{itemize}

    \vspace*{5pt}

    \textbf{\# Input Format \# }
    
    $\{\{\texttt{Question}\}\}$

    \vspace*{5pt}

    \textbf{\# Response Format \#}

    Please return your answer in JSON format:
    \begin{lstlisting}[breaklines=true, basicstyle=\ttfamily, columns=fullflexible]
    {
      "queries": [
        {"angle": "visual", "query": "...", "conf": 0.40},
        {"angle": "...", "query": "...", "conf": 0.60}
      ]
    }
    \end{lstlisting}

    \vspace*{5pt}

    \textbf{\# Example \#}
    
    \textbf{Question:} From 2014 to 2015, which group had the most significant drop in the percentage of households claiming their income was falling behind cost of living, and by how much?
    
    \textbf{Response:}
    \begin{lstlisting}[breaklines=true, basicstyle=\ttfamily, columns=fullflexible]
    {
      "queries": [
        {"angle": "visual",      "query": "falling behind cost of living 2014 2015 chart figure table",              "conf": 0.20},
        {"angle": "entity",      "query": "households group definitions demographic segments income categories",      "conf": 0.35},
        {"angle": "comparative", "query": "largest drop by group 2014 vs 2015 comparison most significant decline",   "conf": 0.30},
        {"angle": "numerical",   "query": "percentage point decrease 2014 2015 by group magnitude difference",       "conf": 0.15}
      ]
    }
    \end{lstlisting}

\end{tcolorbox}
    \label{fig:query_decomposition_prompt}
\end{figure*}

\begin{figure*}[htbp] 
    \centering
\noindent  The prompt for the Surveyor is provided below:
\begin{tcolorbox}[     
    colback=white, 
    colframe=gray,               
    coltitle=white,                    
    colbacktitle=gray,                  
    title=\textbf{Prompt for Global Surveyor} 
]
   
    \textbf{\# Task Description  \# }
    
    Automatically analyzes ALL candidate images that were provided at initialization to give a coarse-grained answer. This agent has been pre-configured with all available images and requires no additional parameters. Use this agent when you need a direct answer based on all available images.

    \vspace*{5pt}

    \textbf{\# Input Format \# }
    
    $\{\{\texttt{Question}\}, \{\texttt{Image}\}\}$

    \vspace*{5pt}

    \textbf{\# Response Format \#}

    Please return your answer in JSON format:
    \begin{lstlisting}[breaklines=true, basicstyle=\ttfamily, columns=fullflexible]
    {
        "coarse_answer": "Your direct answer to the question"
    }
    \end{lstlisting}
\end{tcolorbox}
    \label{fig:survery_prompt}
\end{figure*}

\begin{figure*}[htbp] 
    \centering
    \noindent  The prompt for the Locator is provided below:
    
    \begin{tcolorbox}[     
        colback=white, 
        colframe=gray,               
        coltitle=white,                    
        colbacktitle=gray,                  
        title=\textbf{Prompt for Locator},
        width=\textwidth 
    ]
       
        \textbf{\# Task Description \# }
        
        Analyze all candidate images and select the most relevant images with query-relevant summaries. This agent examines all candidate images, selects relevant images, and provides detailed summaries for the selected images explaining their relevance to the question. Use this agent when you need to identify which images are most relevant and understand their content.
    
        \vspace*{5pt}
    
        \textbf{\# Input Format \# }
        
        $\{\{\texttt{Question}\}, \{\texttt{Image}\}\}$
    
        \vspace*{5pt}
    
        \textbf{\# Response Format \#}
    
        Please return your answer in JSON format:
        \begin{lstlisting}[breaklines=true, basicstyle=\small\ttfamily, columns=fullflexible]
    {
        "choice": List[int],
        "outline_level": "Provide summaries for the selected images, explaining what query-relevant information each contains"
    }
        \end{lstlisting}
    
        \vspace*{5pt}
    
        \textbf{\# Example \#}
    
        Example 1: Question: Who is the person playing a musical instrument in a restaurant?
    
        Response to Example 1:
        \begin{lstlisting}[breaklines=true, basicstyle=\small\ttfamily, columns=fullflexible]
    {
        "choice": [0, 1, 2],
        "outline_level": "Image 0 shows that KFC on Renmin Road has a birthday party on February 3rd with musical entertainment. Image 1 indicates that Shanghai hotels have musical instruments playing during meals, suggesting live music performances. Image 2 shows an invitation letter for a music performance at Qintai Art Museum, which relates to music events."
    }
        \end{lstlisting}
    \end{tcolorbox}
    \label{fig:locator_prompt}
\end{figure*}

\begin{figure*}[htbp] 
    \centering
    \noindent  The prompt for the Extractor is provided below:
    
    \begin{tcolorbox}[     
        colback=white, 
        colframe=gray,               
        coltitle=white,                    
        colbacktitle=gray,                  
        title=\textbf{Prompt for Extractor},
        width=\textwidth 
    ]
       
        \textbf{\# Task Description \# }
        
        Extract detailed information from the images selected by Locator based on specific extraction requirements. This agent performs fine-grained reading of the selected images to extract specific details such as numbers, text, names, locations, dates, times, prices, etc. 
    
        \vspace*{5pt}
    
        \textbf{\# Input Format \# }
        
        $\{\{\texttt{Question}\}, \{\texttt{Image}\}, \{\texttt{Extraction Requirement}\}\}$
    
        \vspace*{5pt}
    
        \textbf{\# Response Format \#}
    
        Please provide the extracted information in JSON format, with image keys as dictionary keys:
        \begin{lstlisting}[breaklines=true, basicstyle=\small\ttfamily, columns=fullflexible]
        {
        "img_0": "Detailed description of information found in image 0, including specific values, text, or other requested details",
        "img_1": "Detailed description of information found in image 1, including specific values, text, or other requested details",
        ...
        }
        \end{lstlisting}
    
        \vspace*{5pt}
    
        \textbf{\# Example1 \#}
    
        Question Context: Where can I find a bookstore that sells rare books?
        
        Extraction Requirement: Extract bookstore names, addresses, and business hours.
    
        Response to Example 1:
        \begin{lstlisting}[breaklines=true, basicstyle=\small\ttfamily, columns=fullflexible]
        {
        "img_1": "Rare Finds Bookstore, located at 123 Main Street, open from 9:00 AM to 6:00 PM",
        "img_5": "Price list showing rare books: Ancient manuscripts $120, Historical documents $200, Rare first editions $350"
        }
        \end{lstlisting}

        \vspace*{5pt}
    
        \textbf{\# Example2 \#}
    
        Question Context: What time is the train departing from Hangzhou to Beijing?
        
        Extraction Requirement: Extract train departure times and destinations.
    
        Response to Example 2:
        \begin{lstlisting}[breaklines=true, basicstyle=\small\ttfamily, columns=fullflexible]
        {
        "img_0": "Train ticket: Hangzhou to Beijing, departure time 14:30, seat number 5A, date March 15",
        "img_2": "Train schedule board showing multiple departures: 08:45 to Shanghai, 14:30 to Beijing, 16:20 to Guangzhou"
        }
        \end{lstlisting}
    \end{tcolorbox}
    \label{fig:Extractor_prompt}
\end{figure*}

\begin{figure*}[htbp] 
    \centering
    \noindent The prompt for the Verifier is provided below:
    
    \begin{tcolorbox}[     
        colback=white, 
        colframe=gray,               
        coltitle=white,                    
        colbacktitle=gray,                  
        title=\textbf{Prompt for Verifier},
        width=\textwidth 
    ]
       
        \textbf{\# Task Description \# }
        
        Evaluates whether the current evidence is sufficient to answer the question accurately. Returns completeness score and gaps. 
    
        \vspace*{5pt}
    
        \textbf{\# Input Format \# }
        
        $\{\{\texttt{Question}\}, \{\texttt{Image}\}, \{\texttt{Evidence Bank}\}\}$
    
        \vspace*{5pt}
    
        \textbf{\# Response Format \#}
    
        Please provide the extracted information in JSON format, with completeness score and gaps as dictionary keys:
        \begin{lstlisting}[breaklines=true, basicstyle=\small\ttfamily, columns=fullflexible]
        {
        "completeness_score": A number from 0-10 (0=insufficient, 10=complete and accurate),
        "gaps": "Description of missing information. Empty string if evidence is complete."
        }
        \end{lstlisting}
    
        \vspace*{5pt}
    
        \textbf{\# Example1 \#}
    
        Question: What time is the train departing from Hangzhou to Beijing?
        
        Evidence Bank: coarse\_answer="14:30", outline\_level="Image 0 shows train ticket with departure 14:30", detail\_level=\{"img\_0": "Train ticket: Hangzhou to Beijing, departure 14:30, seat 5A, date March 15"\}
    
        Response to Example 1:
        \begin{lstlisting}[breaklines=true, basicstyle=\small\ttfamily, columns=fullflexible]
        {
        "completeness_score": 10,
        "gaps": "None"
        }
        \end{lstlisting}

        \vspace*{5pt}
    
        \textbf{\# Example2 \#}
    
        Question: Where can I find a bookstore that sells rare books?
        
        Evidence Bank: coarse\_answer="Rare Finds Bookstore", outline\_level="Image 1 shows 'Rare Finds Bookstore' sign"
    
        Response to Example 2:
        \begin{lstlisting}[breaklines=true, basicstyle=\small\ttfamily, columns=fullflexible]
        {
        "completeness_score": 6,
        "gaps": "Missing address and business hours information"
        }
        \end{lstlisting}
    \end{tcolorbox}
    \label{fig:Verifier_prompt}
\end{figure*}

\end{document}